\pdfoutput=1
\documentclass[sigconf]{acmart}
\acmYear{2026}
\acmConference[DAI '26]
  {The Eighth International Conference on Distributed Artificial Intelligence}
  {November 29--December 2, 2026}
  {Hong Kong, China}
\acmBooktitle{The Eighth International Conference on Distributed Artificial
  Intelligence (DAI '26), November 29--December 2, 2026, Hong Kong, China}
\setcopyright{none}
\copyrightyear{}
\acmDOI{}
\acmISBN{}
\usepackage{array}
\usepackage{booktabs}
\usepackage{placeins}
\usepackage{etoolbox}
\AtBeginEnvironment{thebibliography}{\raggedright}
\usepackage{tikz}
\usetikzlibrary{arrows.meta,calc,positioning,shapes.geometric,shapes.misc,patterns}


\begin{document}

\title[EngineTailor]{EngineTailor: Minimal Complete Inference Engines for
Agent-Driven Optimization}

\makeatletter
\newcounter{inlinenote}
\def\@affiliationnotes{}
\def\@othernotes{}
\def\@inlinenote#1#2{%
  \g@addto@macro\addresses{\@authornotemark}%
  \stepcounter{inlinenote}%
  \edef\@tempa{\noexpand\g@addto@macro\noexpand#1{%
    \noexpand\@textsuperscript{\noexpand\normalfont
      \noexpand\@fnsymbol{\the\c@inlinenote}}\unexpanded{#2}}}%
  \@tempa}
\def\@affiliationname#1{{\def\institution##1{##1}\def\city##1{}%
  \def\country##1{}#1}}
\def\additionalaffiliation#1{%
  \ifx\@affiliationnotes\@empty\else\g@addto@macro\@affiliationnotes{; }\fi
  \@inlinenote\@affiliationnotes{\@affiliationname{#1}}}
\def\authornote#1{\@inlinenote\@othernotes{#1\enspace}}
\def\@authornotes{{\let\@makefnmark\relax\footnotetext{%
  \ifx\@affiliationnotes\@empty\else Also with \@affiliationnotes.\enspace\fi
  \@othernotes\unskip}}}
\makeatother
\author{Haiyan Qin}
\authornote{This work was done during an internship at Infinigence AI.}
\affiliation{\institution{Beihang University}\city{Beijing}\country{China}}
\additionalaffiliation{\institution{Jingjinji National Center of Technology Innovation}\city{Beijing}\country{China}}

\author{Rui Wan}
\affiliation{\institution{Fudan University}\city{Shanghai}\country{China}}
\additionalaffiliation{\institution{Shanghai Innovation Institute}\city{Shanghai}\country{China}}

\author{Jinchen Han}
\affiliation{\institution{Infinigence AI}\city{Shanghai}\country{China}}

\author{Shulin Zeng}
\affiliation{\institution{Infinigence AI}\city{Shanghai}\country{China}}

\author{Guohao Dai}
\affiliation{\institution{Shanghai Jiao Tong University}\city{Shanghai}\country{China}}
\additionalaffiliation{\institution{Infinigence AI}\city{Shanghai}\country{China}}

\author{Yu Wang}
\authornote{Corresponding authors.}
\affiliation{\institution{Tsinghua University}\city{Beijing}\country{China}}
\email{yu-wang@tsinghua.edu.cn}

\author{Wang Kang}
\authornotemark[5]
\affiliation{\institution{Beihang University}\city{Beijing}\country{China}}
\email{wkang@buaa.edu.cn}

\renewcommand{\shortauthors}{Qin et al.}

\begin{abstract}
Production inference frameworks such as vLLM and SGLang span models,
devices, schedulers, and workloads. Their breadth creates a large, coupled
optimization surface: retrieving editable fragments does not give a coding
agent a coherent view of the complete execution path or a change's end-to-end
effects. EngineTailor instead provides each model--device--workload target with a
\emph{minimal but complete} inference engine. It excludes unrelated serving
functionality while retaining the target's complete execution and verification
path. A Manager Agent turns measured bottlenecks into bounded tasks for fresh
Worker Agents, while a protected acceptance plane admits only independently
verified changes that pass correctness and satisfy a predeclared
\textsc{win} or \textsc{enable} contract. Verified changes strengthen the
current target engine, while
reusable implementations and their applicability conditions accumulate in a
shared substrate for re-qualification on later targets.
We evaluate seven targets spanning autoregressive LLM, embodied VLA, and
interactive world-model inference; four are edge-oriented. On a matched
Qwen3-4B/SM100 BF16 task under the same agent configuration, EngineTailor
obtains verified prefill gains after matching token-generation latency,
whereas five direct SGLang attempts yield no admitted improvement. On Jetson AGX Orin, the
accepted strict-BF16 Pi0.5 product reduces end-to-end action latency by 43.0\%
relative to PhyAI (190.3 versus 333.9\,ms). On SM100, re-qualifying a
fixed-contract graph boundary for Qwen raises concurrency-16 (c16) throughput from 3078.6 to
4654.8 tokens/s (51.2\%). These results show that target-scoped engines can
evolve without discarding reusable implementation knowledge. Source code and
artifacts are available at
\url{https://github.com/qhy991/EngineTailor}.
\end{abstract}


\maketitle
\pagestyle{plain}

\begin{figure*}[tp]
  \centering
  \includegraphics[width=\linewidth]{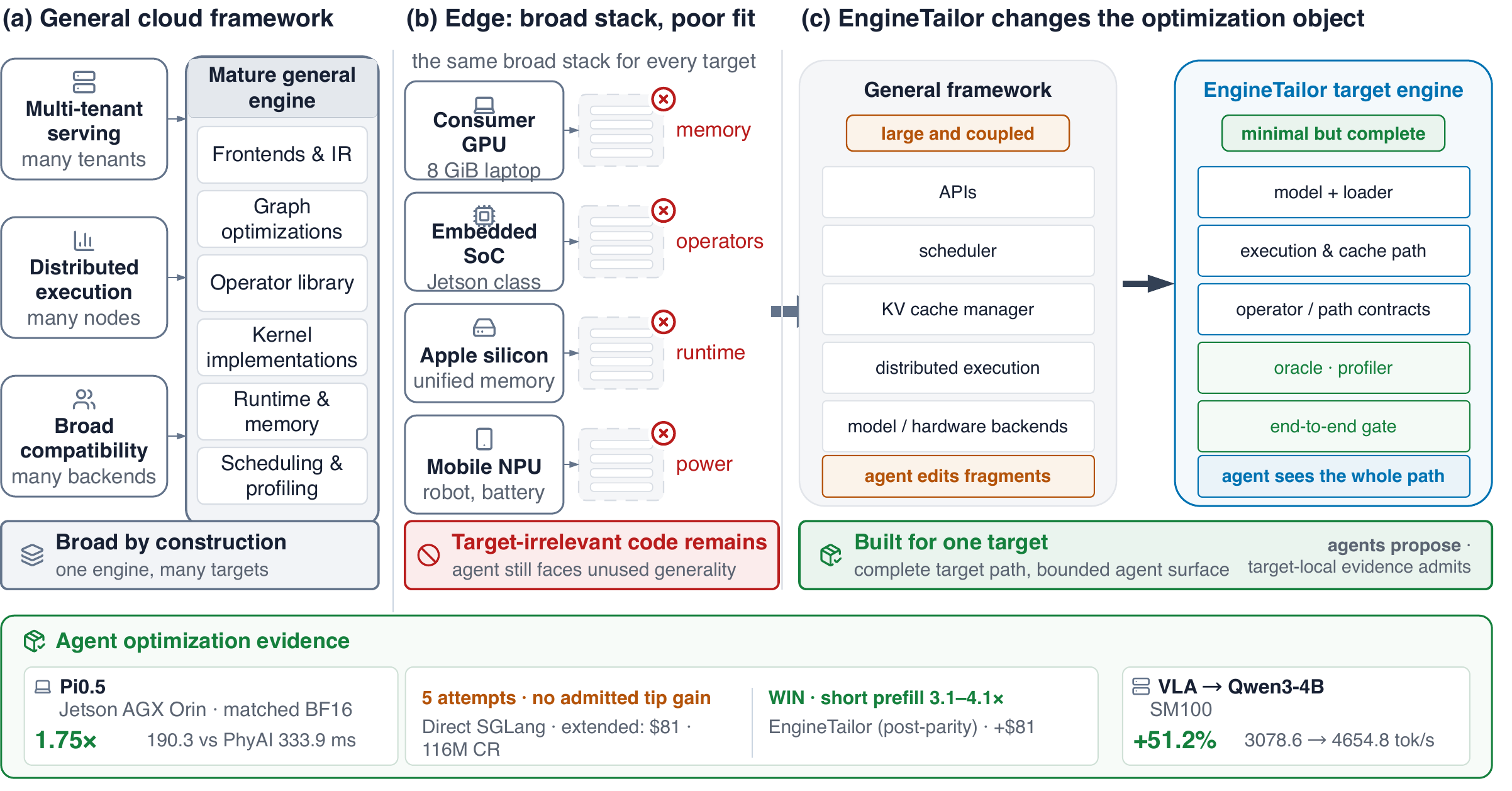}
  \caption{EngineTailor replaces a broad framework with a minimal but complete
  engine for one target.}
  \Description{Three side-by-side panels above a results band. Panel a, General
  cloud framework, shows three cards for multi-tenant serving with many tenants,
  distributed execution with many nodes, and broad compatibility with many
  backends, beside a mature general engine stack listing frontends and
  intermediate representation, graph optimizations, an operator library, kernel
  implementations, runtime and memory, and scheduling and profiling; a strip
  summarizes it as broad by construction, one engine for many targets. Panel b,
  Edge, broad stack, poor fit, shows the same broad stack carried to a consumer
  GPU laptop, an embedded system on chip of the Jetson class, Apple silicon with
  unified memory, and a mobile neural processing unit in a battery-powered
  robot; each target is marked with a failed constraint, namely memory,
  operators, runtime, and power, and a strip states that target-irrelevant code
  remains and the agent still faces unused generality. Panel c contrasts two
  optimization objects. On the left a general framework is labelled large and
  coupled and lists application programming interfaces, a scheduler, a key-value
  cache manager, distributed execution, and model and hardware backends, with
  the agent editing fragments. On the right the EngineTailor target engine is
  labelled minimal but complete and lists model and loader, execution and cache
  path, operator and path contracts, an oracle and profiler, and an end-to-end
  gate, with the complete target path remaining available in the agent-facing
  surface; a strip states that the engine is
  built for one target with a complete target path and a bounded agent surface,
  and that agents propose while target-local evidence admits. The band beneath,
  agent optimization evidence, holds three cards. On the left, Pi0.5 on Jetson
  AGX Orin under a matched strict-BF16 protocol reaches 190.3\,ms versus
  PhyAI's 333.9\,ms, a 1.75-times speedup. In the middle, a matched Qwen3-4B on
  SM100 in BF16 under the same agent configuration is split into two halves:
  five direct SGLang optimization attempts yield no admitted tip gain in the
  extended accounting window, while
  EngineTailor after parity adds short-prefill gains of 3.1 to 4.1 times. On
  the right, a fixed-contract VLA to Qwen3-4B transfer on SM100 raises c16
  throughput from 3078.6 to 4654.8 tokens per second, plus 51.2 percent, after
  fresh target-side re-qualification. Each verdict is named in words as well
  as coloured.}
  \label{fig:motivation}
\end{figure*}

\section{Introduction}

Inference optimization is a system problem. A faster kernel can lose its gain
to layout conversion, graph breaks, cache behavior, host dispatch, or another
phase of the same workload. Production engines such as
vLLM~\cite{kwon2023vllm} and SGLang~\cite{zheng2024sglang} manage these
interactions across many models, devices, and serving modes. That breadth is
essential for a general framework, but it presents a coding agent with a large,
coupled repository whose target path and end-to-end consequences extend beyond
any retrieved fragment. Figure~\ref{fig:motivation} contrasts a broad stack
carried to the edge with EngineTailor's bounded but complete target engine.

Our central research question is: \emph{how should an inference engine be
structured so that a coding agent can reason about and optimize it as a whole?}
Our answer is \emph{one target, one engine}: a \emph{minimal but complete
target engine}. For a registered model--device--workload target,
\emph{minimal} excludes serving generality unrelated to that target, while
\emph{complete} preserves the execution and verification path needed to judge
every candidate by its end-to-end effects.

EngineTailor turns this formulation into a governed evolution process. A
Manager Agent turns measured bottlenecks into bounded tasks. Only target-local
correctness and end-to-end performance evidence may advance the engine.
Verified implementations and their observed conditions remain available to
later targets, which must re-qualify them rather than inherit earlier verdicts.

The evaluation spans seven targets across autoregressive LLM, embodied VLA,
and interactive world-model inference; four are edge-oriented. To isolate the
optimization object from agent capability, the Qwen3-4B/SM100 BF16 audit fixes
the target, task contract, and agent configuration. It measures the resulting
\emph{framework-comprehension tax} through cached-context expenditure and
gate-admitted outcomes (\S\ref{sec:matched_audit}): five direct SGLang attempts yield
no admitted gain, whereas after matching token-generation latency EngineTailor
converts a comparable budget scale into correctness-gated prefill gains while
preserving decode. The
initial Qwen target engine spans 3.3K Python lines, compared with 416K for
SGLang's runtime and 684K for vLLM's package under the same counting rule. On
Jetson AGX Thor, successive verified states reduce Wall-OSS BF16 action latency
by 54.0\%. On SM100, re-qualifying a fixed-contract graph boundary for Qwen
improves c16 throughput by 51.2\%. The same acceptance process also rejects numerically
incorrect candidates and stops correct changes that fail to improve the
complete target workload.

Our contributions are:
\begin{itemize}
  \item \textbf{The minimal-complete engine formulation.} We identify the
  target-scoped optimization object that retains complete execution and evidence
  (\S\ref{sec:substrate}).
  \item \textbf{A governed agent optimization loop.} Bounded agent
  proposals remain separate from protected target-local admission
  (\S\ref{sec:control}, \S\ref{sec:acceptance}).
  \item \textbf{Two-level evolution with executable reuse.}
  Verified target-engine evolution is coupled with conditional reuse across targets
  (\S\ref{sec:evolution}, \S\ref{sec:memory}).
  \item \textbf{A matched audit and cross-platform study.}
  A same-agent-configuration Qwen3-4B/SM100 BF16 audit compares campaign-level
  budget conversion against direct SGLang, while seven LLM, VLA, and
  world-model targets evaluate the formulation and its two evolution loops
  (\S\ref{sec:experiments}).
\end{itemize}

\section{Related Work}
\label{sec:related}

\textbf{Agentic serving and systems optimization.}
VibeServe is the closest work: fresh agents synthesize and validate a complete
runtime from a model, reference, benchmark, and target
hardware~\cite{kamahori2026vibeserve}. We use a pinned VibeSys workspace from
the same project line as a realized serving-system
comparator~\cite{vibesys2026workspacesources}. Its validation is service-level,
whereas EngineTailor admits against a registered model trajectory;
\S\ref{sec:policy} measures one candidate under both policies. EngineTailor
studies repeated evolution
after a minimal but complete target engine has been provided, with admission
and reusable evidence persisting across optimization episodes.

\textbf{Repository agents, bounded search, and verifier integrity.}
SWE-agent and OpenHands expose existing repositories to coding
agents~\cite{yang2024sweagent,wang2024openhands}; SWE-EVO shows that coordinated
multi-file evolution remains harder than isolated issue
resolution~\cite{thai2025sweevo}. General multi-agent frameworks coordinate
cooperating agents through conversation patterns, role programs, or generated
workflows~\cite{wu2023autogen,hong2023metagpt,zhang2024aflow}; EngineTailor
fixes one Manager--Worker decomposition and places final authority in a
protected target-local gate. Performance-oriented systems instead narrow
the object: KernelBench and KernelSkill target individual
kernels~\cite{ouyang2025kernelbench,sun2026kernelskill}, while FlashInfer-Bench
and AgentCompile target deployable kernels or bounded inference
modules~\cite{xing2026flashinferbench,li2026agentcompile}. The AI CUDA Engineer couples
search to an automated verifier~\cite{sakana2025aicuda}, and robust-kbench
demonstrates why evaluator integrity is part of a performance
claim~\cite{lange2025robustkbench}. EngineTailor makes software scope itself a design
choice: its agent-facing object is bounded to one target but retains the full
registered execution and evidence path.

\textbf{Self-evolving agents and executable memory.}
Recent surveys organize self-evolution by the object that changes, including
agent parameters, policies, memory, tools, and
architecture~\cite{gao2025selfevolvingsurvey}. AlphaEvolve evolves executable programs
under automated evaluators~\cite{alphaevolve2025}; G\"odel-machine descendants and the Darwin
G\"odel Machine modify executable agent
code~\cite{schmidhuber2007godel,zhang2025dgm}; SkillRL jointly updates a policy and
retrieved skill bank~\cite{xia2026skillrl}; and MemEvolve adapts both stored
experience and the memory architecture~\cite{zhang2025memevolve}. Engram
instead carries code, logs, results, and a distilled Research Digest through
fresh-agent systems-optimization campaigns~\cite{karimi2026engram}. These
systems evolve programs or agent components, or carry experience across runs.
EngineTailor focuses on the inference-engine lifecycle: one target-local gate
controls the deployed engine state, while runnable implementations and their
observed conditions populate a separate cross-target substrate.
Section~\ref{sec:evolution} defines the two states and their admission rules.

\textbf{Generalist frameworks and target-specific systems.}
vLLM and SGLang are broad serving
stacks~\cite{kwon2023vllm,zheng2024sglang}. Local GGUF deployment systems include
llama.cpp and Ollama; MLX targets Apple Silicon~\cite{llamacpp,ollama,mlx2023}.
FlashInfer and FlashAttention provide reusable attention
kernels~\cite{ye2025flashinfer,dao2022flashattention}. TVM and Ansor search tensor
programs within transformation spaces whose legal operations are bounded before
measurement~\cite{chen2018tvm,zheng2020ansor}. At whole-model compute
scope, MPK and Ada-MK generate target-specific
megakernels~\cite{cheng2025miragepersistent,dong2026adamk}, and AutoMegaKernel guards a
generated schedule with a static checker~\cite{jaber2026automegakernel}. These
routes either provide broad deployment coverage or specialize a bounded compute
region. EngineTailor specializes the complete execution of one registered
model--device--workload target.

\section{Method}

\begin{figure*}[tp]
  \centering
  \includegraphics[width=\linewidth]
    {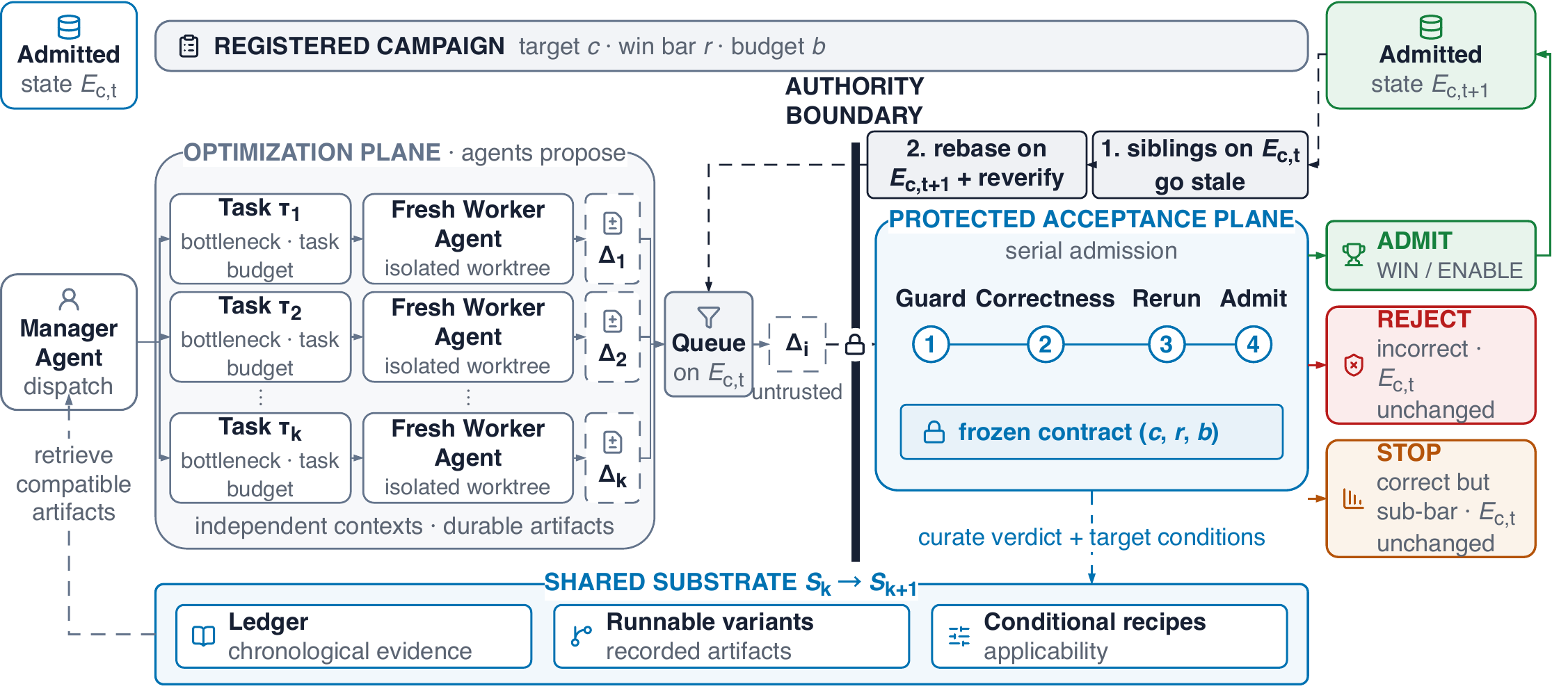}
  \caption{EngineTailor separates parallel agent proposals from serial,
  protected state transitions.}
  \Description{A wide flow diagram read left to right. At the top left the
  admitted engine state E sub c comma t; a wide registered campaign bar naming
  target c, win bar r, and budget b; and at the top right the admitted
  successor E sub c comma t plus one. On the left a Manager Agent fans out to
  three stacked rows, each holding a task scoped to one measured bottleneck,
  then a Fresh Worker Agent on an isolated worktree, then a dashed
  candidate diff; a vertical ellipsis marks that the rows continue. The diffs
  merge into a candidate queue based on E sub c comma t, from which one
  untrusted candidate at a time crosses a heavy black vertical rule marked
  authority boundary into the protected acceptance plane, which admits
  serially. Inside that plane four numbered stages run in series: guard,
  correctness, rerun, and admission, above a box naming the frozen contract c, r,
  b. Three arrows leave the plane to the right: a green WIN that advances E sub
  c comma t to E sub c comma t plus one through an admitted WIN or ENABLE and
  returns up the right margin to the successor state; a red REJECT for
  incorrect candidates; and an orange STOP
  for correct but sub-bar ones. Neither REJECT nor STOP changes the state. Two
  dark outlined boxes above the plane number the consequence of a win: first,
  siblings on E sub c comma t go stale; second, they must be rebased on E sub c
  comma t plus one and reverified. A dashed path connects them and returns to
  the candidate queue. A dashed arrow runs from the acceptance plane down into
  a shared substrate panel labelled S sub k to S sub k plus one, holding a
  ledger, runnable variants, and conditional recipes, curating each verdict
  together with its target conditions; a dashed arrow returns from that panel
  to the Manager Agent, which retrieves compatible artifacts for the next
  task.}
  \label{fig:planes}
\end{figure*}

Figure~\ref{fig:planes} separates three authority planes: agents propose in
the optimization plane, the protected acceptance plane authorizes engine-state
transitions, and the shared-substrate plane preserves executable outcomes. The
method follows a candidate through four questions: what object the agent sees,
how one optimization episode runs, how a candidate is admitted, and how the
result changes later episodes and targets.

\subsection{The Agent-Facing Optimization Object}
\label{sec:frictions}
\label{sec:substrate}

EngineTailor begins by registering one target $c=(m,h,s)$: model $m$, hardware
$h$, and workload and service-level objective $s$. Human governance also fixes a finite campaign budget
$b$, an end-to-end win bar $r$, and a versioned starting engine $E_{c,0}$. The
win bar is the minimum absolute or relative improvement declared before a task
can be recorded as a performance \textsc{win}. A predeclared \textsc{enable}
contract may separately authorize a correctness-verified, non-regressing
default change for a bounded cumulative or safety role; it changes the
registered implementation but is never counted as a \textsc{win}. A
\emph{board} is the frozen benchmark contract
and the measurements collected under it. Each evaluated campaign begins from
a provided target engine whose scope has already been reduced to $c$, making
the complete target execution path, rather than a retrieved fragment, the
Worker's starting optimization object.

The target engine satisfies two complementary conditions. \emph{Minimal}
removes models, backends, serving modes, and distributed machinery unrelated to
$c$. \emph{Complete} retains the model, execution and cache path, swappable
interfaces, profiler, reference, and end-to-end checks for the registered
workload.

Minimality concerns the agent-facing edit surface, not external
dependencies. Protected references, gates, and benchmarks remain callable but
outside the candidate edit surface. At campaign scope, the Manager Agent uses
the current engine state, profile, and prior records to select one bounded
task. The Worker Agent receives that task together with an isolated copy of
the complete target engine. Its objective is local, while the complete target
execution path remains available for inspection.
Table~\ref{tab:minimal-substrate} lists the components that make this object
operational.

\begin{table}[tbp]
  \caption{Operational components of a minimal but complete engine for one
  registered target.}
  \label{tab:minimal-substrate}
  \centering
  \footnotesize
  \setlength{\tabcolsep}{2.5pt}
  \begin{tabular}{@{}>{\raggedright\arraybackslash}p{0.23\columnwidth}>{\raggedright\arraybackslash}p{0.70\columnwidth}@{}}
    \toprule
    Component & Contents and role \\
    \midrule
    Execution path &
    Model and loader, forward or generate entry, cache, scheduling mode, and
    target runtime path; executes the registered workload end to end. \\
    Optimization units &
    Operator and path contracts with inputs, candidate variants, local checking,
    and microbenchmarks; expose measured bottlenecks through stable interfaces. \\
    Evidence path &
    Protected reference, profiler, smoke tests, target gate, and end-to-end
    benchmark; measures headroom and determines admissibility. \\
    Versioned state &
    Base commit, task contract, candidate diff, measurements, and verdict; ties
    every proposed change to the engine state on which it was evaluated. \\
    \bottomrule
  \end{tabular}
\end{table}

Operator units cover individual kernels, while path units cover interactions
such as prefill, decode, cache behavior, and the runtime interface. A Worker can
reuse a community implementation, compose available libraries, or add custom
code through the same contracts. Local checking and microbenchmarks provide a
fast exploratory filter; only complete-engine admission can advance the target
state. Each retained variant records its role, precision, batch regime, context
length, and other conditions, so the engine can select a specialized path and
keep a verified fallback elsewhere.

\subsection{One Optimization Episode}
\label{sec:control}

Starting from the admitted state $E_{c,t}$, one episode moves through the three
planes as follows:

\begin{enumerate}
  \item \textbf{Measure and scope.} The Manager Agent reads the registered
  target, current engine state, profile, recoverable headroom, prior records,
  win bar, and remaining budget. It selects one measured bottleneck and records
  a bounded task contract and base commit.
  \item \textbf{Propose in isolation.} A fresh Worker Agent receives an isolated
  worktree containing the complete target engine and one bounded objective. It
  measures before editing, reuses compatible implementations when available,
  and returns a candidate diff with its local evidence. Several Workers may
  explore different bottlenecks from the same admitted state in parallel.
  \item \textbf{Admit serially.} Candidate diffs enter a queue and cross the
  authority boundary one at a time. The protected plane evaluates each under
  the frozen target contract and independently reruns the evidence.
  \item \textbf{Advance or retain.} A \textsc{win}, or a predeclared
  default-on \textsc{enable}, creates the next engine state. Untested siblings
  based on the previous state become stale and require rebase and full
  re-verification. \textsc{reject} and \textsc{stop} preserve the current
  engine, but their evidence remains available to later tasks.
\end{enumerate}

Humans register the target, win bar, and budget and authorize campaign opening,
closure, or reopening. Within an open campaign, agents perform profiling, task
construction, implementation, and stopping recommendations. A campaign becomes
eligible for closure after reaching its target, exhausting its budget, measuring
remaining headroom below the win bar, or recording three consecutive cycles without a verified win.

\subsection{Protected Target-Local Admission}
\label{sec:acceptance}

Candidate code has no authority over the evidence that admits it. The direct,
deliberately slow reference defines the target behavior, and a guard
freezes the reference functions and judge files. Workers may invoke checks from
their worktrees, but a separate pre-merge path reruns the guard, smoke test,
target gate, and benchmark. This one-way boundary lets candidates consume
verdicts without changing how those verdicts are produced.

\paragraph{Calibrated admission.}
\label{sec:gate}
For autoregressive models, EngineTailor uses teacher forcing as an evaluation
protocol: the protected reference supplies the same prefix to the reference and
candidate at every step, and the gate compares their next-token decisions.
Autoregressive targets require at least 98\% agreement together with semantic
and batched/prefix probes; continuous-output targets use registered error,
shape, and finiteness checks. A candidate that deliberately changes numerical
precision must additionally agree with the accepted reduced-precision path and
hold a quality floor against the reference precision, so a precision change
cannot be admitted on throughput alone. Faithful paths and standing broken controls
calibrate these thresholds.

Only a gate-passing candidate reaches the registered end-to-end benchmark. A
\textsc{win} also clears the win bar and an independent rerun; a
predeclared \textsc{enable} satisfies its non-regression or cumulative-utility
contract without contributing to the WIN count. Both may advance the target
state. \textsc{reject} denotes an integrity or correctness failure, whereas
\textsc{stop} preserves a correct state that does not merit admission.

\subsection{Two-Level Evolution and Verified Reuse}
\label{sec:evolution}

EngineTailor evolves two states over different timescales. The inner loop is the
current target engine $E_{c,t}$: repeated target-local admissions make one
registered model--device--workload engine progressively better. The outer loop
is the shared substrate $S_k$: after each episode, curation links the task and
verdict to an append-only ledger, the runnable candidate variant, and a
conditional recipe. Thus the engine accumulates admitted performance, while
the infrastructure accumulates executable choices and the conditions under
which they should be reconsidered.

Only accepted target-local evidence changes the engine, whereas every curated
verdict can change what later episodes retrieve. Here, $\mathsf{Curate}$ is an
append-only record operation: it stores the candidate, verdict, evidence, and
target conditions for later retrieval without overriding target-local
admission.

Formally, let $\Delta_t$ denote a candidate diff, $\oplus$ its
application to the recorded base state, and
$v_t\in\{\textsc{win},\textsc{enable},\textsc{reject},\textsc{stop}\}$ the
protected verdict:
\[
\begin{aligned}
E'_{c,t} &= E_{c,t}\oplus\Delta_t,\\
v_t &= \mathsf{Eval}_{c}(E'_{c,t}),\\
E_{c,t+1} &=
\begin{cases}
E'_{c,t}, & v_t\in\{\textsc{win},\textsc{enable}\},\\
E_{c,t}, & v_t\in\{\textsc{reject},\textsc{stop}\},
\end{cases}\\
S_{k+1} &= \mathsf{Curate}(S_k,c,\Delta_t,v_t).
\end{aligned}
\]

A substrate record makes this update executable. It binds the source and target
configuration, engine base, candidate artifact and entry point, gate version
and control result, local and end-to-end measurements, protected verdict,
rollback, and any condition that would justify reopening the direction. The
record therefore preserves both an implementation that can run again and the
evidence needed to decide whether it is relevant to a later target.

A verified reusable implementation can enter the variant portfolio, while a
mechanistic \textsc{stop} or \textsc{reject} can narrow a
recipe's preconditions or record a counterexample. The Manager Agent retrieves
records compatible with the current target, bottleneck, measured headroom, and
gate version. An unsuccessful direction is suppressed while those conditions
remain unchanged, but becomes eligible again when a relevant condition changes.
Retrieved evidence changes which proposal is tried next; it does not change the
target's admission rule.

For a new target $c'$, EngineTailor starts from its own provided minimal but
complete engine $E_{c',0}$. The Manager Agent retrieves a compatible mechanism
and runnable implementation from $S_k$, and a Worker adapts it to the new target.
The new target's guard, correctness gate, and task contract then return
\textsc{reject}, \textsc{stop}, \textsc{enable}, or \textsc{win}. The resulting
evidence updates the substrate with the new conditions, while only a new
target-local admitted verdict advances $E_{c',t}$. This is the connection between the two forms
of self-evolution: later tasks can begin from accumulated capability without
inheriting another target's verdict.

\section{Experiments}
\label{sec:experiments}

We ask three questions: how observed agent budget conversion differs between
optimization objects under matched conditions; whether target specialists
reach useful endpoints across registered boards; and what the design
contributes: whether complete admission changes which candidates advance,
how gains accumulate within a target, and what survives re-qualification when
an implementation moves to another target.

\subsection{Experimental Setup and Optimization Objects}
\label{sec:setup}

We evaluate seven model--device--workload targets spanning autoregressive LLMs,
embodied VLAs, and an interactive world model. The edge targets are
Wall-OSS-0.5/Thor~\cite{yu2026walloss}, Pi0.5/Orin~\cite{pi2025pi05},
Qwen3.5-2B/RTX~5070~\cite{qwen35-2026}, and Qwen3-0.6B/M1
Pro~\cite{qwen3-2025}. Qwen3-4B~\cite{qwen3-2025},
Llama-3.1-8B~\cite{grattafiori2024llama3}, and the Oasis-500M interactive
world model~\cite{etched2024oasis} run on SM100. Their registered workloads
cover batch-one generation, continuous serving, VLA action chunks, and a
200-frame interactive rollout; Appendix~\ref{app:protocol} gives the complete
contract registered for each target.

All reported runs use Codex \texttt{gpt-5.6-sol} at \texttt{xhigh} under a
protocol registered before dispatch. The authors prepared the versioned
starting engines with coding-agent assistance before the campaigns, outside
the campaign budgets (Appendix~\ref{app:protocol}). Campaigns dispatch
bounded units under that protocol: the Thor campaign, for example, dispatched
24 units, 23 of which completed, and produced three independently verified
merges. Within a campaign, agents perform profiling, task selection,
implementation, evaluation, and stopping. Costs use a frozen price table, and
CR denotes cache-read input tokens.

Campaigns run in one of three registered operating modes. The Qwen3-4B/SM100
campaign is strict: after the frozen $E_{c,0}$, every engine change is
agent-authored (\S\ref{sec:matched_audit}). The Wall-OSS/Thor campaign is
agent-run: agents executed the dispatched units while a human owner directed
registration, continuation, and the campaign pause. The remaining five
campaigns are assisted, with humans and agents working inside the same
protocol. The acceptance plane is mode-blind: every admitted change, in every
mode, cleared the same target gate and independent rerun.

Peer boards use SGLang for stable language-serving comparisons, the pinned
VibeSys artifact where registered~\cite{vibesys2026workspacesources}, PhyAI on
Orin~\cite{phyai2026}, and a frozen Ollama comparison on M1. When a stable peer
is unavailable, the headline board is self-relative; the RTX peer measurement
remains in the appendix because its 8k arm is unstable. Each target couples its
end-to-end metric with a protected reference, a target-specific gate, and a
standing broken control. Reported values are medians over archived samples,
and the archived spread is small against the reported margins: the c16
opposite-order pairs land at $+50.25\%$/$+51.70\%$, the refreshed RTX medians
carry coefficients of variation of 4.4\%/1.7\%, and the five timed PhyAI
samples span 333.33--334.90\,ms. Appendix Table~\ref{tab:e2e_results} gives
the full boards.

We also record the agent-facing software surface of the matched Qwen objects.
At frozen commits, the initial target engine contains 28 Python files and 3.3K
physical lines against SGLang's 416K and vLLM's 684K under the same counting
rule (Table~\ref{tab:surface-summary}), while retaining generation, cache,
profiling, benchmarking, and gate paths.
This comparison describes software surface; by itself, it does not measure
agent comprehension or search efficiency.

\begin{table}[tbp]
  \centering
  \caption{Agent-facing Python surface at frozen commits; values are rounded.}
  \label{tab:surface-summary}
  \scriptsize
  \setlength{\tabcolsep}{3pt}
  \begin{tabular}{@{}>{\raggedright\arraybackslash}p{0.50\columnwidth}rr@{}}
    \toprule
    \textbf{System / counted tree} & \textbf{Files} & \textbf{Lines} \\
    \midrule
    EngineTailor / initial Qwen target & 28 & 3.3K \\
    SGLang / runtime & 1.0K & 416K \\
    vLLM / package & 1.8K & 684K \\
    \bottomrule
  \end{tabular}
\end{table}

\subsection{Matched Optimization-Object Audit}
\label{sec:autonomy}
\label{sec:matched_audit}

Table~\ref{tab:matched-object-audit} compares observed budget conversion on
the same Qwen3-4B/SM100 BF16 target,
objective, and registered task under a fixed agent configuration. After the
frozen $E_{c,0}$, all reported Qwen engine changes are agent-authored, and each
cycle records its task contract, candidate diff, measurements, gate,
verification, and verdict.

\begin{table}[tbp]
  \centering
  \caption{Parity-anchored Qwen3-4B/SM100 same-task campaign audit. Both arms
  use Codex \texttt{gpt-5.6-sol} at \texttt{xhigh}; CR is cache-read input;
  PP and TG denote prefill and token generation.}
  \label{tab:matched-object-audit}
  \footnotesize
  \setlength{\tabcolsep}{3pt}
  \begin{tabular}{@{}>{\raggedright\arraybackslash}p{0.23\columnwidth}>{\raggedright\arraybackslash}p{0.25\columnwidth}>{\raggedright\arraybackslash}p{0.43\columnwidth}@{}}
    \toprule
    \textbf{Object / window} & \textbf{Agent expenditure} &
    \textbf{Gate-admitted outcome} \\
    \midrule
    SGLang, strict-BF16 slice & \$54.87; 74.77M CR (96.9\%) &
    PP/TG change $<1\%$; no tip-level WIN \\
    SGLang, extended cutoff & approximately \$81.39; 115.76M CR &
    covers all five direct attempts; no admitted tip-level gain \\
    EngineTailor, post-parity BF16 & approximately \$81 &
    Short PP $3.10$--$4.05\times$; long PP $1.25$--$1.41\times$; TG preserved \\
    \bottomrule
  \end{tabular}
\end{table}

Five direct SGLang attempts produce no admitted tip-level gain (no change to
the enabled default path). To compare the
two objects from a common performance footing rather than from their own
starting engines, we anchor the audit at the first same-machine batch-one
comparison in which EngineTailor is no longer behind SGLang on token
generation, 2.79 versus approximately 3.09\,ms/token. From that anchor,
EngineTailor's next approximately \$81 admits $3.10$--$4.05\times$
short-prefill gains at 64/128/256 input tokens and $1.25$--$1.41\times$
long-prefill gains at 2048/4096 input tokens over each task's pre-change
EngineTailor path, while preserving decode. The
approximately \$69 spent reaching the anchor is recoverable and
excluded, and the audit measures observed campaign-level budget conversion
between optimization objects, not an isolated surface-size effect;
Appendix~\ref{app:protocol} records the accounting boundary. At comparable
budget under the same agent configuration, spending therefore converted into an
admitted end-to-end gain on the target engine and into none on the framework.

\subsection{Cross-Target Endpoints and Boundaries}
\label{sec:cloud}

We next examine whether target-scoped engines reach useful endpoints across the
seven registered targets. Table~\ref{tab:headline_results} groups external-peer
and self-relative boards. For Qwen, the campaign audit in
\S\ref{sec:matched_audit}, the deployment endpoint here, and the
conditional-reuse diagnostic in \S\ref{sec:memory} are separate registered
boards with distinct denominators. Appendix Table~\ref{tab:e2e_results} gives the per-depth
measurements.

\begin{table}[tbp]
  \centering
  \scriptsize
  \setlength{\tabcolsep}{3pt}
  \caption{Current target-local endpoints, grouped by baseline type; effects
  are not compared across boards.}
  \label{tab:headline_results}
  \begin{tabular}{@{}>{\raggedright\arraybackslash}p{0.31\columnwidth}>{\raggedright\arraybackslash}p{0.41\columnwidth}>{\raggedright\arraybackslash}p{0.21\columnwidth}@{}}
    \toprule
    \textbf{Target} & \textbf{Final vs. comparator / initial} &
    \textbf{Effect} \\
    \midrule
    \multicolumn{3}{@{}l}{\textit{External-peer boards}} \\
    Pi0.5 / Orin & 190.3 vs. PhyAI 333.9 ms &
    $43.0\%$ lower \\
    Qwen3-4B / SM100 & 2.669 vs. SGLang 3.533 ms/tok at 4k &
    $24.4\%$ lower \\
    Llama-3.1-8B / SM100 & 3465 vs. SGLang 4171 tok/s at c16 &
    $16.9\%$ below SGLang \\
    \midrule
    \multicolumn{3}{@{}l}{\textit{Self-relative boards}} \\
    Wall-OSS-0.5 / Thor & 280.7 vs. initial 610.3 ms &
    $2.17\times$ faster \\
    Qwen3-0.6B / M1 & 7.173 vs. pre-admission 8.303 ms/tok &
    $13.6\%$ lower \\
    Qwen3.5-2B / RTX~5070 & 68.4 vs. initial 32.8 tok/s at 4k &
    $2.09\times$ higher \\
    Oasis-500M / SM100 & 26.31 vs. initial 612.5 ms/frame &
    $23.3\times$ faster \\
    \bottomrule
  \end{tabular}
\end{table}

\paragraph{External-peer boards.}
\label{sec:external-peer-results}
On Orin, the accepted BF16 Pi0.5 path reaches 190.3\,ms from 866.2\,ms,
43.0\% below PhyAI's 333.9\,ms. On Qwen3-4B/SM100, the registered board lowers
BF16 decode latency by 24.4\%/23.1\% and BF16 prefill latency by
6.2\%/5.1\% at 4k/8k versus SGLang, and FP8 prefill by 6.6\%/4.1\%. On Llama-3.1-8B,
the same substrate reaches batch-one decode parity and lowers prefill by
2.8\%/6.3\%, while c16 serving on the mixed/BF16 deployment board reaches
3464.8 versus SGLang's 4171.2\,tok/s.

\paragraph{Self-relative boards.}
\label{sec:self-relative-results}
Successive admissions move the Wall-OSS/Thor BF16 engine from 610.3 to
280.7\,ms ($-54.0\%$, $2.17\times$). On RTX~5070, BF16 generation rises
from 32.8 to 68.4/62.7\,tok/s at 4k/8k. On M1 Pro, the current W8 endpoint is
7.173\,ms/token; Appendix Table~\ref{tab:e2e_results} retains the separate
frozen Ollama board. On Oasis-500M, saturated-frame latency falls from 612.5 to
26.31\,ms/frame ($23.3\times$) under the complete 200-frame rollout gate. All
seven registered endpoints improve on their own board; the single headline
board on which a peer remains ahead is Llama at concurrency 16.

\subsection{Design Diagnostics}

\begin{figure*}[tp]
  \centering
  \includegraphics[width=\textwidth]
    {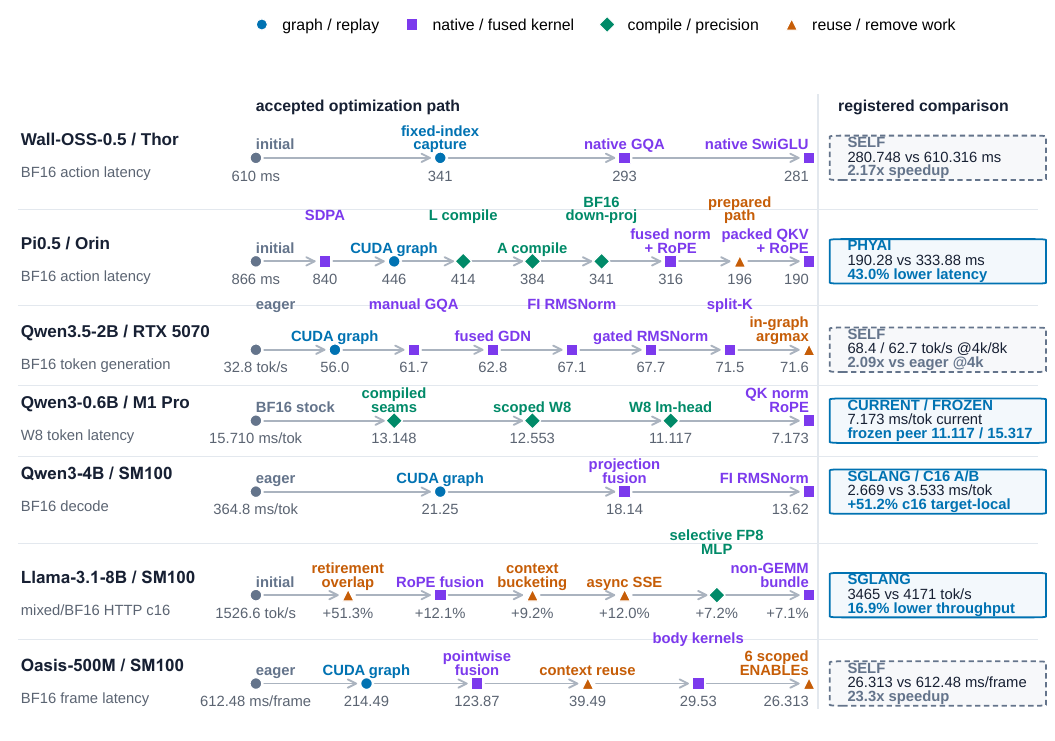}
  \caption{Selected admitted optimization paths and registered target comparisons.}
  \Description{Seven horizontal lanes show selected changes that advance each
  target engine. Marker color and shape group graph or replay, native or fused
  kernels, compilation or precision, and reuse or removal of repeated work.
  A separate card compares the registered endpoint with PhyAI, Ollama, or
  SGLang when a stable deployment board exists; Thor, RTX 5070, and Oasis use
  their initial engine because no stable matched peer board is available.}
  \label{fig:verified_trajectories}
\end{figure*}

The boards above report what the process produced; we now ask which parts of
the design account for it. We begin with the cases that tested the design by
disagreement, where narrow and complete-engine evidence pointed in opposite
directions, then examine the correctness plane those verdicts rely on, how
admitted changes accumulate within a target
(Figure~\ref{fig:verified_trajectories}), and what an admitted implementation
carries to a later target.

\paragraph{Completeness stress test.}
Representative archived cases show that narrower and complete-engine evidence
can disagree in both directions. Two candidates that passed narrower checks
regressed the complete board: a Pi0.5 attention swap measured $1.497\times$
locally yet added 56.7\,ms end to end, while an RTX single-pass decode passed its
registered 98\% correctness gate at 126/128 token steps but reduced 4k/8k
throughput from 67.7/62.4 to 60.2/52.4\,tok/s. Other candidates exposed integration or coverage failures
absent from the narrower evidence: FlashInfer decode attention measured $1.2$--$1.8\times$
at the operator level but was not graph-capturable, and an Oasis candidate that was 5.30\% faster in same-session
A/B failed the complete 200-frame rollout gate. A Llama path improved step
latency from 4.901 to 4.382\,ms but produced only a 0.6--2.0\% serving gain,
below its registered bar. Conversely, a Qwen RMSNorm candidate that failed a
local filter passed the complete board and reduced decode latency from 17.96
to 13.61\,ms/token. The complete evidence path therefore decides which
candidates advance, not only whether they are correct. Appendix
Table~\ref{tab:operator_results} gives additional tabulated verdicts.

\paragraph{Admission calibration.}
\label{sec:calibration}

\begin{figure}[tbp]
  \centering
\definecolor{GateBlue}{RGB}{0,114,178}
\definecolor{GateGreen}{RGB}{21,128,61}
\definecolor{GateRed}{RGB}{185,28,28}
\definecolor{GateNeutral}{RGB}{100,116,139}

\begin{tikzpicture}[
  x=0.0336cm,
  y=0.615cm,
  font=\sffamily\fontsize{7.2}{8.4}\selectfont,
  faithful/.style={draw=GateBlue,fill=GateBlue,line width=.8pt},
  broken/.style={draw=GateRed,line width=1.0pt},
  guide/.style={draw=black!12,line width=.35pt},
  value/.style={anchor=west,font=\sffamily\bfseries\fontsize{7.2}{8.4}\selectfont}
]
  \fill[GateBlue!5]  (-2,3.14) rectangle (101.5,5.72);
  \fill[GateRed!5]   (-2,1.16) rectangle (101.5,2.94);
  \node[anchor=west,text=GateBlue,font=\sffamily\bfseries\fontsize{7.2}{8.4}\selectfont]
    at (-1,5.46) {gate-admitted};
  \node[anchor=west,text=GateRed,font=\sffamily\bfseries\fontsize{7.2}{8.4}\selectfont]
    at (-1,2.68) {standing broken controls};

  \foreach \x in {0,25,50,75,100} {
    \draw[guide] (\x,0.74) -- (\x,5.72);
    \node[anchor=north,text=black!65] at (\x,0.58) {\x};
  }
  \draw[black!55,line width=.5pt] (0,0.74) -- (101.5,0.74);
  \node[anchor=north,font=\sffamily\fontsize{7.2}{8.4}\selectfont] at (50,-0.02)
    {teacher-forced top-1 agreement (\%)};

  \draw[GateGreen,dashed,line width=1.0pt] (98,0.74) -- (98,6.10);
  \node[anchor=south east,text=GateGreen,font=\sffamily\bfseries\fontsize{7.2}{8.4}\selectfont]
    at (99.5,6.14) {98\% acceptance bar};

  \node[anchor=east] at (-3,4.88) {SM100 faithful};
  \draw[GateBlue,line width=1.2pt] (0,4.88) -- (99.2,4.88);
  \fill[faithful] (99.2,4.88) circle (2.3pt);
  \node[value,text=GateBlue] at (105,4.88) {99.2\% \,(127/128)};

  \node[anchor=east] at (-3,4.16) {SM100 reference};
  \draw[GateBlue,line width=1.2pt] (0,4.16) -- (100,4.16);
  \fill[faithful] (100,4.16) circle (2.3pt);
  \node[value,text=GateBlue] at (105,4.16) {100\%\phantom{.0} \,(128/128)};

  \node[anchor=east] at (-3,3.44) {RTX 5070 faithful};
  \draw[GateBlue,line width=1.2pt] (0,3.44) -- (98.4,3.44);
  \fill[faithful] (98.4,3.44) circle (2.3pt);
  \node[value,text=GateBlue] at (105,3.44) {98.4\% \,(126/128)};

  \node[anchor=east] at (-3,2.22) {SM100 broken};
  \draw[GateRed,line width=1.0pt] (0,2.22) -- (14.1,2.22);
  \node[broken,inner sep=1.4pt,font=\sffamily\bfseries\fontsize{7.2}{8.4}\selectfont,text=GateRed]
    at (14.1,2.22) {$\times$};
  \node[value,text=GateRed] at (105,2.22) {14.1\% \,(18/128)};

  \node[anchor=east] at (-3,1.50) {RTX 5070 broken};
  \draw[GateRed,line width=1.0pt] (0,1.50) -- (84.4,1.50);
  \node[broken,inner sep=1.4pt,font=\sffamily\bfseries\fontsize{7.2}{8.4}\selectfont,text=GateRed]
    at (84.4,1.50) {$\times$};
  \node[value,text=GateRed] at (105,1.50) {84.4\% \,(108/128)};
\end{tikzpicture}
  \caption{Teacher-forced acceptance separates faithful paths from standing
  controls.}
  \Description{A horizontal agreement plot shows SM100 and RTX 5070 faithful
  paths above the 98 percent bar and their standing broken controls below it.
  Counts out of 128 decisions accompany each row.}
  \label{fig:calibration}
\end{figure}
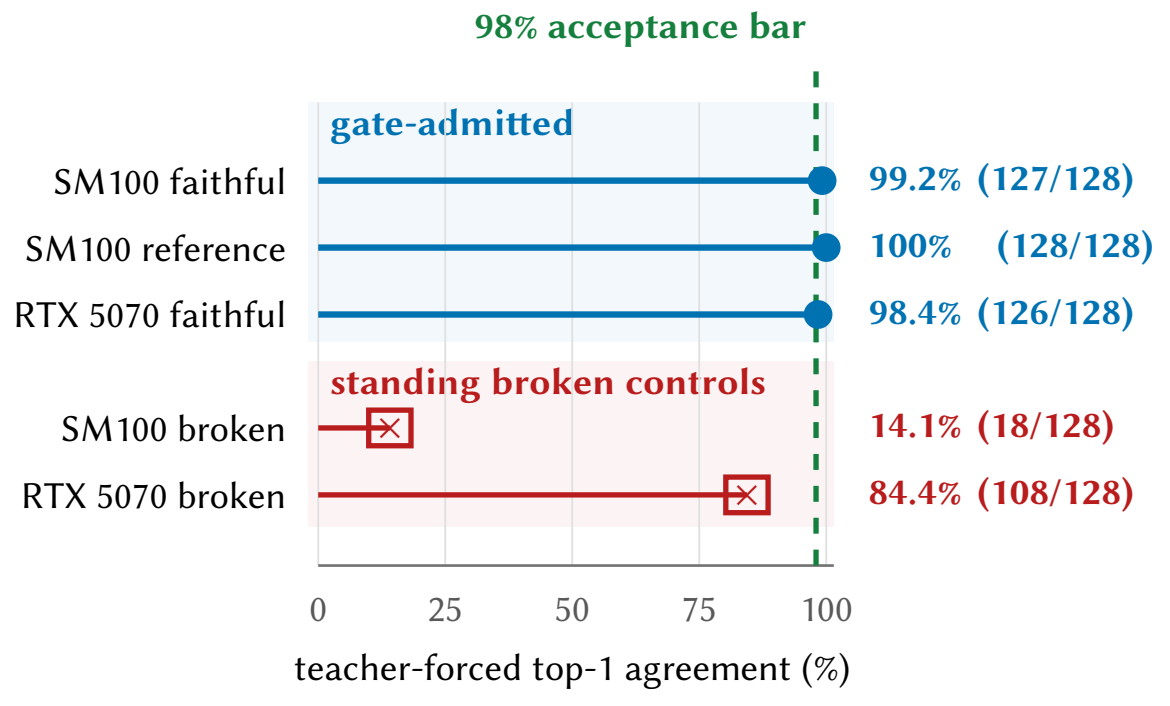

The acceptance plane must detect incorrect candidates before performance is
considered. At the registered 98\% threshold, faithful rows score 98.4\%
or higher, while standing broken controls score 11.9--14.3\% on SM100 and
84--86\% on RTX (Figure~\ref{fig:calibration});
Appendix~\ref{app:acceptance-details} expands the probes.

The same authority is specialized to each output type. Pi0.5's continuous
action path clears its registered relative-error bar with the expected finite
shape, while its broken control fails, and Oasis applies its gate to the
complete 200-frame rollout. On every registered target the thresholds
therefore separate faithful paths from their standing controls, including the
RTX control that sits closest to the acceptance bar.

\paragraph{Acceptance policy is not interchangeable.}
\label{sec:policy}
The same candidate can be admissible under one project's checker and
inadmissible under another's. A VibeSys from-scratch Llama-3.1-8B/SM100 tip
enables reduced precision for MLP, attention, and the language-model head by
default and passes its own service checker on sentinel, known-answer, and
determinism probes. Teacher-forced against that tip's own forced-BF16 path, it
agrees on 124 of 128 next-token decisions and 123 of 128 at $B{=}4$, below the
registered bar, while the forced-BF16 arm agrees on all 128 and the negative
control is convicted at 1 of 128. Under our policy the configuration would ship
env-gated and default-off rather than as a default precision. The frozen Qwen
comparison is unaffected, since that arm runs BF16 with its optional
reduced-precision path disabled. The two policies answer different questions: a
service checker asks whether an endpoint behaves plausibly, the target gate
whether a candidate preserves a registered model trajectory, so acceptance
policy and not engine structure alone decides what ships.

\paragraph{Accumulation.}
Admitted changes follow a recurring order across targets.
Figure~\ref{fig:verified_trajectories} names selected admitted changes that
build each target engine and keeps the registered comparator separate from the
historical path. The paths distinguish bar-clearing WINs from scoped ENABLEs
and show order and accumulation; they do not treat differences between
checkpoints as matched causal effects. Across targets, graph capture or replay
removes the first large structural bottleneck; native kernels, compilation,
precision choices, and removal of repeated work then compound the accepted
state. Matched A/B measurements link the largest structural gains to Thor
capture ($-43.8\%$) and Oasis context reuse ($-68.1\%$).

\paragraph{Conditional reuse.}
\label{sec:memory}

Evolution within a target binds every performance decision to one model,
device, workload, and gate. Across targets, the substrate makes later searches
cumulative by retaining runnable variants and verified conditions.

The Wall-OSS campaigns connect the two levels across devices. Fixed-index
capture, native GQA, and native SwiGLU first enter the shared substrate from the
SM100 campaign and are then re-qualified under Thor's SM110 gate and end-to-end
board. Across workloads, a fixed graph-boundary implementation raises
Qwen3-4B/SM100 c16 throughput from 3078.6 to 4654.8\,tok/s ($+51.2\%$).

Cross-model reuse of Llama/SM100 context-bucketed attention raises Qwen c16
throughput from 2803.6 to 3191.3\,tok/s ($+13.8\%$). Cross-device reuse of a
SM100 stack raises RTX prefill throughput by 21.8\%/48.5\% at 4k/8k.
Unsuccessful retargeting records the condition that blocked reuse. Reuse
therefore provides a tested starting point for later searches without
transferring a verdict: every gain above was re-earned under the receiving
target's own gate.

\subsection{Scope and Threats to Validity}
\label{sec:limits}
``Minimal'' refers to the target-relevant agent-facing surface; external
libraries remain trusted dependencies. Thor and Oasis use self-relative
anchors, and RTX remains target-local. Pi0.5 and Wall-OSS verify action-tensor
fidelity for one chunk; task-level robot behavior is outside this study.
Transfer is measured between provided target engines rather than from automatic
framework extraction.

The optimization-object audit is one target-local
contrast; the seven-target results establish coverage of the formulation, not
repetition of that contrast. CR is a repeated-context proxy, and costs are
price-indexed estimates. Appendix~\ref{app:protocol} gives the full protocols and
release boundary.

\FloatBarrier
\section{Discussion and Conclusion}
EngineTailor makes a minimal but complete target engine the object of
agent-driven optimization: bounded contracts focus proposals, while a protected
gate that candidates cannot edit retains end-to-end authority over engine
state.

Two observations extend beyond the endpoint measurements. First, under matched
target, task, and agent conditions, the two optimization objects show
different budget conversion. From a common performance footing and at comparable
campaign scales, five direct SGLang attempts admitted no tip-level gain while
EngineTailor admitted $3.10$--$4.05\times$ short-prefill improvement. Coding-agent effectiveness is therefore conditioned
by both the agent configuration and the optimization object it faces. Second,
narrow evidence is insufficient as the sole admission authority.
Representative cases disagreed with the complete board in both directions:
locally attractive candidates regressed or failed deployment conditions, while
a Qwen RMSNorm candidate rejected by a local filter reduced decode latency by
24\% on the complete board at that checkpoint.

More broadly, parallel proposals from a shared admitted state, serial
adjudication behind a one-way authority boundary, and append-only curation of
verdicts and conditions suggest a reusable governance pattern for agents that
modify shared executable state. This design still requires a prepared starting
engine for each target, outside the measured campaign budget. EngineTailor
assumes that the shared substrate amortizes this preparation as verified
implementations accumulate. Reuse observed in the seven-target study is
consistent with that assumption, but does not establish whether retrieval and
re-qualification remain effective as the substrate and target set grow.

\section*{Generative AI Use Disclosure}
Generative AI tools supported literature and repository analysis, experimental
workflow design, kernel-code generation and analysis, experiment orchestration,
and language editing. The authors defined the research questions and protocols,
verified admitted code and reported measurements against protected tests and
archived artifacts, and take full responsibility for the content and integrity
of this work.

\begin{acks}
This work was supported by the Beijing Natural Science Foundation (L2607008).
\end{acks}

\bibliographystyle{ACM-Reference-Format}
\bibliography{refs}

\appendix
\setlength{\dbltextfloatsep}{12pt plus 2pt minus 2pt}
\setlength{\dblfloatsep}{8pt plus 1pt minus 1pt}
\section{Pre-Registered Protocol and Disclosure}\label{app:protocol}
\begin{table*}[tp]
  \centering
  \scriptsize
  \renewcommand{\arraystretch}{0.95}
  \setlength{\tabcolsep}{2.4pt}
  \caption{Expanded per-depth deployment boards.}
  \label{tab:e2e_results}
  \begin{tabular}{@{}>{\raggedright\arraybackslash}p{0.13\textwidth}>{\raggedright\arraybackslash}p{0.19\textwidth}>{\raggedright\arraybackslash}p{0.07\textwidth}>{\raggedright\arraybackslash}p{0.17\textwidth}>{\raggedright\arraybackslash}p{0.12\textwidth}>{\raggedright\arraybackslash}p{0.19\textwidth}>{\raggedright\arraybackslash}p{0.06\textwidth}@{}}
    \toprule
    \textbf{Target} & \textbf{Metric} & \textbf{Prec.} & \textbf{Input} &
    \textbf{EngineTailor} & \textbf{Peer / baseline} & \textbf{Ratio} \\
    \midrule
    Qwen3-4B / SM100 & TG latency & BF16 & 4k & 2.669 ms/tok &
    SGLang 3.533 ms/tok & 1.324$\times$ \\
    Qwen3-4B / SM100 & TG latency & BF16 & 8k & 2.804 ms/tok &
    SGLang 3.645 ms/tok & 1.300$\times$ \\
    Qwen3-4B / SM100 & PP latency & BF16 & 4k & 27.89 ms &
    SGLang 29.73 ms & 1.066$\times$ \\
    Qwen3-4B / SM100 & PP latency & BF16 & 8k & 61.41 ms &
    SGLang 64.68 ms & 1.053$\times$ \\
    Qwen3-4B / SM100 & PP latency & FP8 & 4k & 23.02 ms &
    SGLang 24.66 ms & 1.071$\times$ \\
    Qwen3-4B / SM100 & PP latency & FP8 & 8k & 52.33 ms &
    SGLang 54.55 ms & 1.042$\times$ \\
    Qwen3-4B / SM100 & PP latency & FP8 & B16$\times$512 &
    ${\sim}33.7$ ms & SGLang ${\sim}39.5$ ms & ${\sim}1.17\times$ \\
    Qwen3-4B / SM100 & current HTTP throughput & BF16 & c16 &
    4654.812 tok/s & pre-admission 3078.559 tok/s & 1.512$\times$ \\
    Qwen3-4B / SM100 & historical HTTP throughput & BF16 & c16 &
    1796.4 tok/s & VibeSys 1609.2 tok/s & 1.116$\times$ \\
    Qwen3-4B / SM100 & TG latency & BF16 & 4k &
    2.669 ms/tok & VibeSys direct 6.986 ms/tok & 2.618$\times$ \\
    \midrule
    Llama-3.1-8B / SM100 & TG latency & BF16 & 4k &
    4.4018 ms/tok & SGLang 4.3853 ms/tok & 0.996$\times$ \\
    Llama-3.1-8B / SM100 & TG latency & BF16 & 8k &
    4.5062 ms/tok & SGLang 4.4899 ms/tok & 0.996$\times$ \\
    Llama-3.1-8B / SM100 & PP latency & BF16 & 4k &
    43.97 ms & SGLang 45.25 ms & 1.029$\times$ \\
    Llama-3.1-8B / SM100 & PP latency & BF16 & 8k &
    94.47 ms & SGLang 100.85 ms & 1.068$\times$ \\
    Llama-3.1-8B / SM100 & HTTP throughput & mixed/BF16 & c16 &
    3464.81 tok/s & SGLang 4171.20 tok/s & 0.8306$\times$ \\
    Llama-3.1-8B / SM100 & HTTP throughput & BF16 & c16 &
    2786.36 tok/s & VibeSys BF16 2394.53 tok/s & 1.164$\times$ \\
    Oasis-500M / SM100 & current saturated-frame latency & BF16 &
    200 frames/W32/DDIM10 & 26.313 ms/frame & initial 612.48 ms/frame & 23.277$\times$ \\
    Oasis-500M / SM100 & last WIN-qualified state & BF16 &
    200 frames/W32/DDIM10 & 29.53 ms/frame & initial 612.48 ms/frame & 20.741$\times$ \\
    \midrule
    Qwen3.5-2B / RTX~5070 & TG throughput & BF16 & 4k &
    68.4 tok/s & eager 32.8 tok/s & 2.085$\times$ \\
    Qwen3.5-2B / RTX~5070 & TG throughput & BF16 & 8k &
    62.7 tok/s & -- (peer not admitted) & -- \\
    Qwen3.5-2B / RTX~5070 & TG throughput & mixed & 4k &
    76.9 tok/s & SGLang FP8 80.58 tok/s & 0.954$\times$ \\
    Qwen3.5-2B / RTX~5070 & PP throughput & BF16 & 4k &
    7890 tok/s & SGLang 8562 tok/s & 0.922$\times$ \\
    Qwen3.5-2B / RTX~5070 & PP throughput & FP8 & 4k &
    11025 tok/s & SGLang 16134 tok/s & 0.683$\times$ \\
    Qwen3.5-2B / RTX~5070 & TG throughput & BF16/F16 & 4k &
    76.5 tok/s & Ollama 69.6 tok/s & 1.099$\times$ \\
    Qwen3.5-2B / RTX~5070 & TG throughput & BF16/F16 & 8k &
    75.3 tok/s & Ollama 68.2 tok/s & 1.104$\times$ \\
    Qwen3.5-2B / RTX~5070 & PP throughput & BF16 & 4k &
    4976 tok/s & pre-transfer 4087 tok/s & 1.218$\times$ \\
    Qwen3.5-2B / RTX~5070 & PP throughput & BF16 & 8k &
    4969 tok/s & pre-transfer 3346 tok/s & 1.485$\times$ \\
    \midrule
    Wall-OSS-0.5 / AGX Thor & action latency & BF16 &
    two $448^2$ cams, H=32, flow=10 & 280.748 ms &
    Phase-0 610.316 ms & 2.174$\times$ \\
    Pi0.5 / Jetson Orin & action latency & BF16 & M=968, flow=10 &
    190.28 ms & PhyAI 333.88 ms & 1.755$\times$ \\
    Qwen3-0.6B / M1 Pro & current wall-token latency & W8 & 512+128 &
    7.173 ms/tok & pre-admission 8.303 ms/tok & 1.158$\times$ \\
    Qwen3-0.6B / M1 Pro & frozen deployment board & W8 / Q8\_0 & 512+128 &
    11.117 ms/tok & Ollama 15.317 ms/tok & 1.378$\times$ \\
    \bottomrule
  \end{tabular}
\end{table*}
\begin{table*}[tp]
  \centering
  \scriptsize
  \renewcommand{\arraystretch}{0.95}
  \setlength{\tabcolsep}{2pt}
  \caption{Additional operator and path verdicts.}
  \label{tab:operator_results}
  \begin{tabular}{@{}>{\raggedright\arraybackslash}p{0.12\textwidth}>{\raggedright\arraybackslash}p{0.18\textwidth}>{\raggedright\arraybackslash}p{0.14\textwidth}>{\raggedright\arraybackslash}p{0.24\textwidth}>{\raggedright\arraybackslash}p{0.12\textwidth}>{\raggedright\arraybackslash}p{0.06\textwidth}@{}}
    \toprule
    \textbf{Target} & \textbf{Change} & \textbf{Local evidence} &
    \textbf{Engine-level result} & \textbf{Gate / control} & \textbf{Verdict} \\
    \midrule
    Qwen/SM100 & FlashInfer decode attention & $1.2$--$1.8\times$ op-level &
    Not graph-capturable & -- & REJECT \\
    Qwen/SM100 & FlashInfer RMSNorm & Local filter failed &
    $17.96\rightarrow13.61$ ms/token & $99.2\%$ & WIN \\
    Qwen/SM100 & GEMV quantization & -- & Roofline bound capped the ideal gain at $\sim1.3\times$ & -- & STOP \\
    Llama/SM100 & block-paged KV path & $4.901\rightarrow4.382$ ms/step &
    only $0.6$--$2.0\%$ serving gain & $99.4\%$ ($1527/1536$) & STOP \\
    RTX~5070 & SM100 prefill transplant & -- &
    $4087\rightarrow4976$ / $3346\rightarrow4969$ tok/s &
    $126/128$; control $84.4\%$ & WIN \\
    RTX~5070 & MLP gate/up FP8 & -- &
    $1.065\times$ / $1.088\times$ prefill & $98.4$--$99.2\%$ & WIN \\
    RTX~5070 & FlashInfer GDN chunk core & -- &
    $1.35\times$ / $1.45\times$ prefill & $127/128$ & WIN \\
    Wall-OSS/Thor & SM110 FP8 portfolio & legal consumers measured &
    no accepted FP8 default & PASS / PASS & STOP \\
    Pi0.5/Orin & fused AdaRMSNorm + RoPE & $1.70\times$ / $2.49\times$ op-level &
    $347.55\rightarrow316.00$ ms & relerr $3.906{\times}10^{-3}$ & WIN \\
    Pi0.5/Orin & FlashInfer attention swap & attention $1.497\times$ &
    e2e regresses by 56.7 ms & PASS & STOP \\
    \bottomrule
  \end{tabular}
\end{table*}
The original paper protocol fixed two primary configurations before dispatch:
(i) Qwen3.5-2B on an RTX~5070, batch one, and (ii) Qwen3-4B on one SM100,
batch one. Both use the same protected technical route: a frozen eager oracle,
reuse-first search, a calibrated target gate, independent verification, and
append-only recording. Each task also registers a win bar, a
wall-clock/candidate budget, and a headroom citation. The stopping rule is task
completion, a written performance bound below the win bar, budget exhaustion,
or three consecutive cycles without a verified win.

The recovered Qwen3-4B/SM100 optimization-object audit fixes one SM100, the BF16
accepted path, prefill and token-generation objectives, and the same registered
agent configuration in both arms. The fixed cost rows retain
their original session allowlists and chronological cutoffs. The complete record
contains five direct SGLang optimization attempts on this same target; none
changes the enabled tip. The EngineTailor post-parity window begins at the first clear
same-machine batch-one TG comparison, 2.79 versus approximately
3.09\,ms/token, and excludes the approximately \$69 recoverable
parity-acquisition cost. Dollar values equal recorded token usage times the
frozen model-price table and are estimates, not bills. CR denotes cache-read
input tokens; output tokens are retained separately as generated-work
diagnostics. The final evidence bundle records the session allowlists,
chronological cutoff, price snapshot, task contracts, gates, and STOP sequence.

The authors prepared all seven versioned starting engines with coding-agent
assistance before their registered campaigns. The Qwen engine predates the
original protocol. Historical person-hour and token costs for starting-engine
construction were not recorded uniformly and are excluded from the reported
optimization-cost accounting for every target.

The Oasis-500M/SM100 protocol fixes BF16, batch one, $360{\times}640$
frames, a 32-frame window, ten DDIM steps, and a 200-frame rollout. Its primary
metric is median-of-five CUDA-Event saturated-frame latency after two warmups;
the rollout gate fixes relative $L_2\leq8\times10^{-3}$ and cosine $\geq0.999$
with worst-frame, decoded-frame, and rotary-control checks.

The Wall-OSS/Thor protocol fixes one SM110 device, batch one,
two $448{\times}448$ cameras, 32 actions of dimension 26, ten flow steps, TF32
off, and synchronized median-of-five action-chunk latency. Its continuous gate
requires relative $L_2\leq8\times10^{-3}$, cosine $\geq0.999$, finite
$(1,32,26)$ output, and detection of the negative control. The campaign
dispatched 24 unique units, 23 of which completed, and produced three
independently verified merges; the final accepted implementation is BF16.

For Qwen3-4B on SM100, CUDA graph, projection fusion, and RMSNorm
re-qualification move the historical accepted path from 364.8 to 13.62\,ms per
token. The current registered board reaches 2.669\,ms per token versus
SGLang's 3.533\,ms per token at 4k.

The consumer-GPU result uses Qwen3.5-2B on an RTX~5070 at batch one. Its
2026-08-02 BF16 refresh covers 4k and 8k prompts with 128 generated tokens,
one discarded warmup and five retained samples in each AB and BA order. The
EngineTailor medians are 68.4 and 62.7\,tok/s; SGLang's 8k order split is
58.09/14.93\,tok/s, so no refreshed cross-depth peer ratio is admitted. The
Ollama deployment board and transfer-composition experiment remain separate
registered measurements.

The acceptance-policy comparison in \S\ref{sec:policy} audits an archived
VibeSys from-scratch Llama-3.1-8B/SM100 tip. Its service checker records
sentinel, known-answer, and determinism probes with no local reference model,
teacher-forced trajectory, or standing numerical control. The teacher-forced
counts were collected in process against that same tip's forced-BF16 path at
$n{=}128$, with the negative control convicted at 1 of 128; the relative
perplexity change of the reduced-precision arm, $+0.904\%$, would clear a
quality proxy, but the gate remains the sole merge authority. The frozen Qwen
VibeSys board is separate from this audit and runs with its optional reduced-precision
path disabled.

The Orin comparison fixes the box, weights, M=968 prefix, BF16 precision, flow
schedule, timer, and sample counts for both engines. Cross-platform results use
the registered board for each target, not a single universal
configuration; most boards emphasize batch one, and Llama at concurrency 16
shows that continuous batching remains target dependent. The artifact records
the source snapshots, task contracts, raw samples, candidate diffs, verdicts,
and gate configurations.

\section{Evaluation Boards}\label{app:boards}

Ratios use comparator/EngineTailor for latency and EngineTailor/comparator for
throughput, so values above one favor EngineTailor. TG and PP denote token
generation (decode) and prefill; c16 denotes 16 concurrent requests; B16$\times$512 is
a batch of 16 sequences with 512 input tokens each. RTX uses its fresh transfer
board and Orin the matched CUDA-Event remeasurement. M1 separates the current
native-W8 self-relative A/B from the frozen native-W8 versus GGUF-Q8\_0 board;
Qwen similarly separates the current c16 A/B from its historical VibeSys board.
Dashes mark boards without an admitted peer ratio.

\subsection{Additional Acceptance Evidence}\label{app:acceptance-details}
The Llama gate records 127/128 agreement and passes batched, semantic, and
negative-control probes. Across the SM100 distinct-B16, distinct-B64, and
batched-prefill B16$\times$128 probes, faithful paths record 99.2\%, 99.4\%,
and 99.0\% agreement, while broken controls record 12.5\%, 14.3\%, and
11.9\%. Pi0.5 records $3.906\times10^{-3}$ relative error at an
$8\times10^{-3}$ bar, and Oasis matches the reference over the full 200-frame
rollout. These probes determine the verdicts reported in
Table~\ref{tab:operator_results}.

\clearpage
\end{document}